\documentclass[10pt,twocolumn,letterpaper]{article}

\usepackage{iccv}
\usepackage{times}
\usepackage{epsfig}
\usepackage{graphicx}
\usepackage{amsmath}
\usepackage{amssymb}
\usepackage{booktabs}
\usepackage[pagebackref=true,breaklinks=true,letterpaper=true,colorlinks,bookmarks=false]{hyperref}

\iccvfinalcopy % *** Uncomment this line for the final submission

\ificcvfinal\fi

\begin{document}

%%%%%%%%% TITLE
\title{Learning Foresightful Dense Visual Affordance \\ for Deformable Object Manipulation}

\author{\textbf{Ruihai Wu}$^{1,3,4}$\thanks{Equal contribution, order determined by coin flip.} \quad  \textbf{Chuanruo Ning}$^{2,1}$\footnotemark[1] \quad \textbf{Hao Dong}$ ^{1,3,4}$\thanks{Corresponding author}\\
$^1$CFCS, School of CS, PKU \quad % Peking University \quad
$^2$School of EECS, PKU \quad %Peking University\quad\\
$^3$BAAI \quad
\\$^4$National Key Laboratory for Multimedia Information Processing, School of CS, PKU \quad\\
% {\tt\normalsize \{wuruihai,hao.dong\}@pku.edu.cn},
% {\tt\normalsize chuanruo@stu.pku.edu.cn}
% \\
% \\ \\
% \url{https://hyperplane-lab.github.io/vat-mart}
}

% \author{Ruihai Wu\thanks{Equal contribution, order determined by coin flip.} \qquad \qquad \qquad Chuanruo Ning\footnotemark[1] \qquad \qquad \qquad Hao Dong\thanks{Corresponding author}\\
% CFCS, School of CS, PKU \qquad School of EECS, PKU \qquad CFCS, School of CS, PKU \\
% { BAAI \quad\quad\quad\quad\quad\quad\quad\quad\quad\quad\quad\quad\quad\quad\quad\quad\quad\quad\quad\quad\quad BAAI} \\
% { National Key Laboratory for Multimedia Information Processing \quad National Key Laboratory for Multimedia Information Processing}
%  \\ %Peking University\quad \\
% {\tt\normalsize wuruihai@pku.edu.cn \qquad chuanruo@stu.pku.edu.cn \quad hao.dong@pku.edu.cn}
% % \\
% % \\ \\
% % \url{https://hyperplane-lab.github.io/vat-mart}
% }

% \author{Ruihai Wu\\
% \and

\maketitle
\ificcvfinal\thispagestyle{empty}\fi

%%%%%%%%% ABSTRACT

\begin{abstract}

Understanding and manipulating deformable objects (\emph{e.g.}, ropes and fabrics) is an essential yet challenging task with broad applications.
Difficulties come from complex states and dynamics, diverse configurations and high-dimensional action space of deformable objects. 
Besides, the manipulation tasks usually require multiple steps to accomplish, and greedy policies may easily lead to local optimal states.
Existing studies usually tackle this problem using reinforcement learning or imitating expert demonstrations, with limitations in modeling complex states or requiring hand-crafted expert policies.
In this paper, we study deformable object manipulation using dense visual affordance, with generalization towards diverse states, and propose a novel kind of foresightful dense affordance, which avoids local optima by estimating states' values for long-term manipulation. 
We propose a framework for learning this representation, with novel designs such as multi-stage stable learning and efficient self-supervised data collection without experts.
Experiments demonstrate the superiority of our proposed foresightful dense affordance.
Project page: \href{https://hyperplane-lab.github.io/DeformableAffordance}{https://hyperplane-lab.github.io/DeformableAffordance}

\end{abstract}

% \vspace{-3mm}

\vspace{-1.5mm}
\section{Introduction}
\label{sec:introduction}
\vspace{-1.5mm}

% trim={left button right up}
\begin{figure}[t]
  \centering
  \includegraphics[width=\linewidth, trim={0cm, 0cm, 0cm, 0cm}, clip]{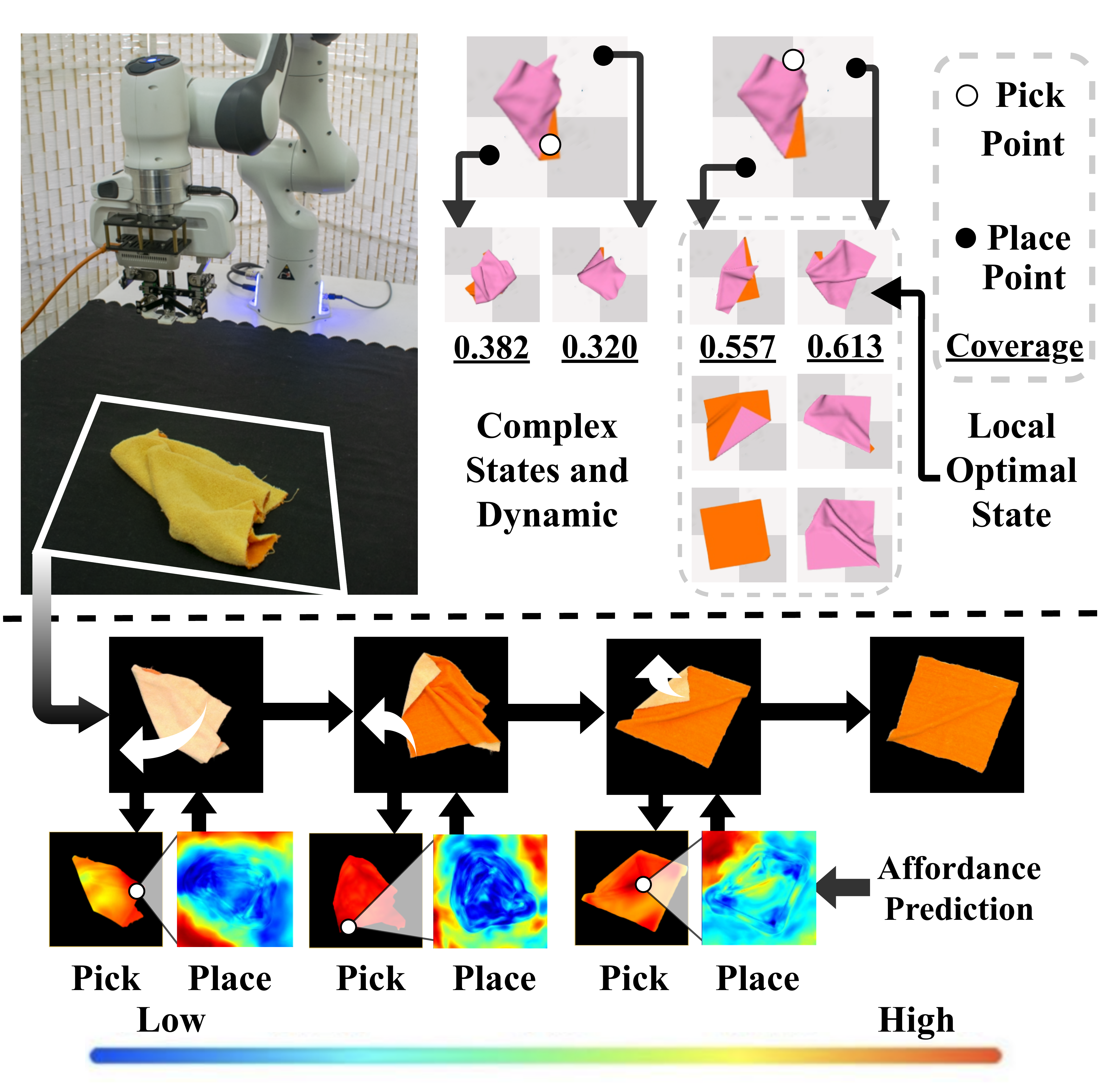}
  \caption{
  \textbf{Deformable Object Manipulation} has many difficulties. 1) It requires \textbf{multiple steps} to complete. 2) Most actions can hardly facilitate tasks, for the \textbf{exceptionally complex states and dynamics}.
  3) Many \textbf{local optimal states} are temporarily closer to the target, but making following actions harder to coordinate for the whole task.
  We propose to learn \textbf{Foresightful Dense Visual Affordance} aware of future actions to avoid local optima for deformable object manipulation, with real-world implementations.
  }
  \label{fig_teaser}
  \vspace{-3mm}
\end{figure}

Many kinds of deformable objects, such as ropes and fabrics, exist everywhere in our daily life.
Perceiving and manipulating deformable objects plays a significant role and paves the way for future home-assistant robots.

Unlike rigid or articulated objects, due to the complex dynamics, high-dimensional and nearly infinite degrees of freedom, large action space, and severe self-occlusion, deformable objects pose much more challenges to manipulate.
Moreover, unlike tasks for rigid objects (like grasping) or articulated objects (like pushing a door) that require one or only a few steps to accomplish, deformable object manipulation tasks usually require many steps to accomplish, laying much focus on relationships and influences between actions in a sequence, as an action leading to local optimal states may not eventually complete the task.

Specifically, as shown in Figure~\ref{fig_teaser}, unfolding crumpled cloth requires a sequence of actions (pick-and-place).
Because of the exceptionally complex states and dynamics, and large action space, most actions fail to facilitate the task.
Moreover, although some cloth in local optimal states temporarily have larger coverage areas than others, following actions face difficulties in smoothly completing the task.

Proposed by Gibson~\cite{gibson1977theory} and aimed at providing indicative information for agents to execute actions (\textit{e.g.}, elementary actions such as picking and pulling) and thus facilitating downstream tasks, visual affordance is arousing much attention in vision and robotics. 
Recent works have demonstrated its efficiency in a large range of tasks like grasping~\cite{mandikal2020graff, montesano2009learning, corona2020ganhand, mia2017affordance, kokic2020learning, zeng2018robotic}, manipulating articulated objects~\cite{mo2021where2act, wu2022vatmart, wang2021adaafford, act_the_part} and assisting robots in a scene~\cite{interaction-exploration, nagarajan2020ego, goff2019building}.
Among them, point-level dense affordance~\cite{wu2022vatmart, mo2021where2act, mo2021o2oafford, wang2021adaafford} learns whether an action on each point of the object could facilitate the task. Compared with Reinforcement Learning (RL) approaches, dense affordance is stably supervised and has better generalization ability towards objects with diverse shapes.%is stably learned with stable ground truth supervision on each point.

The above dense affordance is suitable for representing deformable objects with complex states, capable of indicating whether diverse actions could help complete the task.

While most previous works only study dense affordance for short-term manipulation on rigid~\cite{zhao2022dualafford} or articulated objects~\cite{mo2021where2act, wu2022vatmart},
to tackle the local optima problem in multi-step manipulation,
we move a step towards equipping dense affordance with foresightfulness for future states.

Inspired by Dynamic Programming with Bellman Equation~\cite{bellman1966dynamic} and Q-Learning~\cite{watkins1992q},
estimating a state's \textbf{`value'} (expected return in the long term, instead of only the current performance) for future actions to coordinate and eventually complete the task will help avoid local optimal states and boost the smoothness and quickness of multi-step manipulation.
Dense affordance is suitable for such `value' estimation,
because
such `value' requires understanding and aggregating a large number of diverse following actions on complex states and their corresponding results. 

With such state `value's (instead of only the current performance) in supervisions, dense affordance would gain foresightfulness for the future.

We propose to learn dense visual affordance for manipulating deformable objects,
and further estimate state `value's by aggregating such affordance to avoid local optima and smoothly accomplish multi-step tasks.
As shown in Figure~\ref{fig_teaser} (Down), the task can be accomplished smoothly using our proposed dense affordance in the real world.
To learn such representations, we propose a novel framework generic to diverse tasks with many novel designs, such as the stage-by-stage stable training and \textit{Fold to Unfold} efficient multi-stage data collection.
Thus the proposed affordance could be learned stably, efficiently, and self-supervisedly without hand-crafted policies for different tasks.
Experiments on representative benchmark tasks demonstrate our framework's impressive performance.

\vspace{1mm}
In summary, our contributions are:
\begin{itemize}
    \vspace{-1mm}
    \item We propose to use dense visual affordance for manipulating deformable objects with complex states and dynamics, using such representation to estimate state `value's for future actions to avoid local optima and smoothly accomplish multi-step manipulation tasks;
    \vspace{0mm}
    \item We propose a self-supervised framework with novel designs such as multi-stage training and efficient data collection to learn the proposed affordance stably; 
    \vspace{-3mm}
    \item Qualitative and quantitative results on representative benchmarks and real-world experiments demonstrate the superiority of our proposed dense visual affordance and learning framework for deformable objects.
\end{itemize}

\vspace{-1.5mm}
\section{Related Work}
\vspace{-1.5mm}

% \vspace{-1mm}
\subsection{Perceive and Manipulate Deformable Objects}

% Compared with manipulating deformable objects,
For deformable object perception,
previous works learn and leverage appearance and geometric information ~\cite{nascimento2019geobit, moreno2011deformation,simo2015dali}, softness ~\cite{cellini2013visual}, visual-tactile sensation~\cite{cui2020grasp}, shape completion~\cite{chi2021garmentnets}, emotional perception ~\cite{lee2018anthropomorphic}, optimal perception~\cite{cuiral2020rgb} and the continuity of perception~\cite{martinez2019continuous,florence2018dense,kumari2016haptic}.
The manipulation is challenging for its high complexity.
Typical methods utilize Reinforcement Learning (RL) or Imitation Learning. State-based RL methods~\cite{jangir2020dynamic} consume object states to perform objects manipulation, while
others~\cite{wu2019learning, lee2020learning} utilize visual RL to manipulate rope and cloth with limited states or targets. Transporter~\cite{zeng2021transporter, seita2021learning} exploits imitation learning to perform different tasks, requiring human-designed expert policy and demonstrations for each task, limiting the generalization over tasks and objects. 
Vision-based methods~\cite{ganapathi2021learning, jia2018manipulating} use visual feedback or correspondence to perform manipulations.
Flow-based methods~\cite{shen2022acid, weng2022fabricflownet} learn forward dynamics for manipulation, requiring much time in planning.
Instead of learning flow-based dynamics or from demonstrations, our work builds a bridge between perception and manipulation, taking advantage of the generalization ability of affordance and estimating state `value' for efficient planning.

% \vspace{-2mm}
\subsection{Visual Affordance for Robotic Manipulation}

Learning visual affordance~\cite{gibson1977theory} aims to learn representations of objects or scenes that indicate possible ways for robots to interact and complete tasks.
Many recent works learn affordance for grasping~\cite{mandikal2020graff, montesano2009learning, corona2020ganhand, mia2017affordance, kokic2020learning, zeng2018robotic}, articulated object manipulation~\cite{mo2021where2act, wu2022vatmart, wang2021adaafford, act_the_part, geng2022end}, object-object interaction~\cite{mo2021o2oafford}, collaboration~\cite{zhao2022dualafford} and interaction in a scene~\cite{interaction-exploration, nagarajan2020ego, goff2019building}.
Most tasks can be achieved in a single step (\emph{e.g.}, grasping objects, pulling drawers), and learned affordance only contains actionable information for single-step manipulation.
However, deformable objects with complex states and dynamics require many steps to manipulate.
We move a step towards proposing dense affordance for deformable objects, which not only indicates complex states and dynamics but also takes the subsequent actions of a certain state into consideration for long-term tasks to avoid local optima.
FlingBot~\cite{ha2021flingbot} learns affordance for unfolding cloth with flinging actions, while our proposed affordance is more generic for large action space and diverse tasks.

\vspace{-1.5mm}
\section{Problem Formulation}
\vspace{-1.5mm}

\label{sec_formulation}

In this section, we describe how we formulate learning \textbf{policy for multi-step deformable object manipulation} into learning \textbf{policies for picking and placing}.

Following current benchmarks \textit{DeformableRavens}~\cite{seita2021learning} and \textit{SoftGym}~\cite{corl2020softgym}, we formulate the problem as learning a policy $\pi$ that outputs robot action $a_t$, given the visual observation $o_t$ (denoted as $o$) at time $t$. We use pick-and-place as primitive action, \emph{i.e.}, $a_t = (a_{pick},\ a_{place})$, with $a_{pick}$ and $a_{place}$ denoting picker pose for picking and placing. As explained in ~\cite{seita2021learning}, the picker can complete the tasks without rotation, so $a_{pick}$ and $a_{place}$ can be denoted as picking point $p_{pick}$ and placing point $p_{place}$.

The composite of $p_{pick}$ and $p_{place}$ makes up a large combinatorial action space hard for a network to learn directly and simultaneously.
For the underlying nature that $p_{place}$ highly depends on $p_{pick}$, 
we follow~\cite{wu2019learning} and disentangle learning the composite of $p_{pick}$ and $p_{place}$ into respectively learning $p_{pick}$ and $p_{place | p_{pick}}$ (denoted as $p_{place}$).

Therefore, we formulate the problem into learning picking and placing policies. In Method Section~\ref{sec:method}, we describe how we learn the policies using \textbf{foresightful dense visual affordance} with each state's \textbf{`value'} to avoid local optima.

\section{Method}
\label{sec:method}

\subsection{Overview}

\vspace{-1mm}

\begin{figure}[h]
  \centering
%   \fbox{\rule[-.5cm]{0cm}{4cm} \rule[-.5cm]{4cm}{0cm}}
  \includegraphics[width=\linewidth]{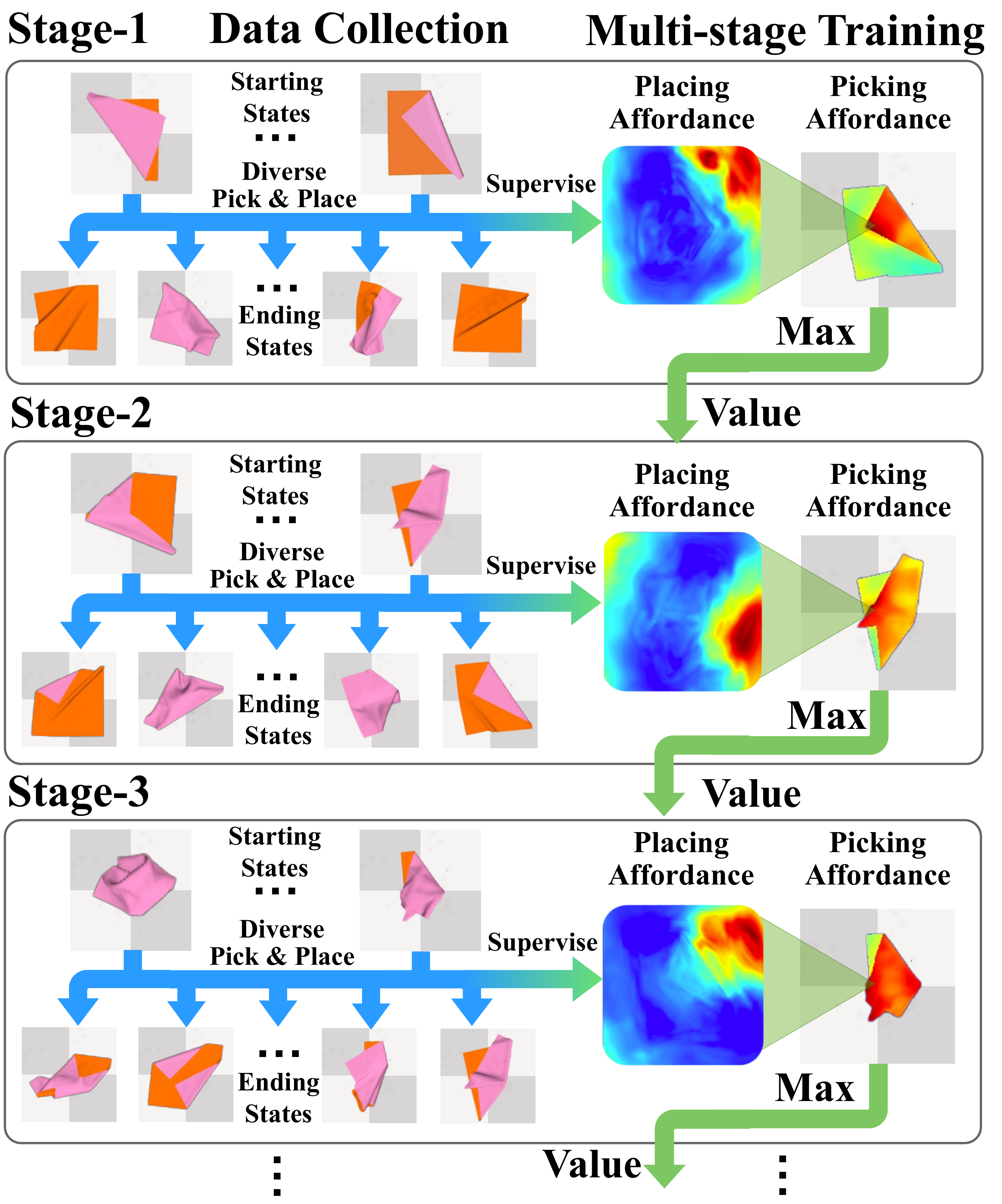}
  \caption{\textbf{Our proposed framework} learns dense picking and placing affordance for deformable object manipulation (\emph{e.g.}, \textbf{Unfolding Cloth}).
  We collect multi-stage interaction data efficiently (Left) and learn proposed affordance stably in a multi-stage schema (Right) in the reversed task accomplishment order, from states close to the target to complex states.
  }
  \vspace{-3.5mm}
  \label{fig_framework}
\end{figure}

As shown in Figure~\ref{fig_framework},
our framework is composed of two main parts:
(1) we propose to use dense affordance representing manipulation policy (\ref{sec:aff_policy}), estimate state `value' using dense affordance and incorporate `value' into dense affordance to avoid local optima for multi-step manipulation (\ref{sec:value}), break the picking-placing dependency cycle and stably learn affordance stage by stage (\ref{sec:learn_aff});
(2) to tackle the difficulty in collecting multi-stage and successful interactions, we propose a method (named \textit{Fold to Unfold}) generic to many tasks to efficiently collect data in the reversed task completion order (\emph{e.g.}, collecting unfolding data by folding cloth) (\ref{sec:data}). 
Besides,
we propose Integrated Systematic Training to further integrate the proposed affordance into a whole system (\ref{sec:ist}).
Finally, 
we describe network architectures and loss function (\ref{sec:network}).

% trim={<left> <lower> <right> <upper>}

\vspace{-1.5mm}
\subsection{Dense Visual Affordance Representing Policy}
\label{sec:aff_policy}
\vspace{-1mm}

This section introduces how to use dense visual affordance to represent manipulation policy.
For simplicity, we first discuss affordance for a greedy policy.

Described in Section~\ref{sec_formulation}, we formulate the manipulation problem into learning picking and placing policies, \emph{i.e.}, the picking point $p_{pick}$ and placing point $p_{place}$ given the observation $o$ (with the size of $m\times n$ points).
As the action space is all points on the object for picking, and all points in the space for placing, 
it comes naturally to use dense picking affordance map $A_{o}^{pick}$ (size $m\times n$) indicating how picking each point will facilitate the task, and dense placing affordance map $A_{o|{p_{pick}}}^{place}$ (size $m\times n$) indicating how placing the picking point $p_{pick}$ on each point will facilitate the task.

\begin{figure}[h]
  % \vspace{-2mm}
  \centering
  \includegraphics[width=\linewidth,  trim={0cm, 0cm, 0cm, 0cm}, clip]{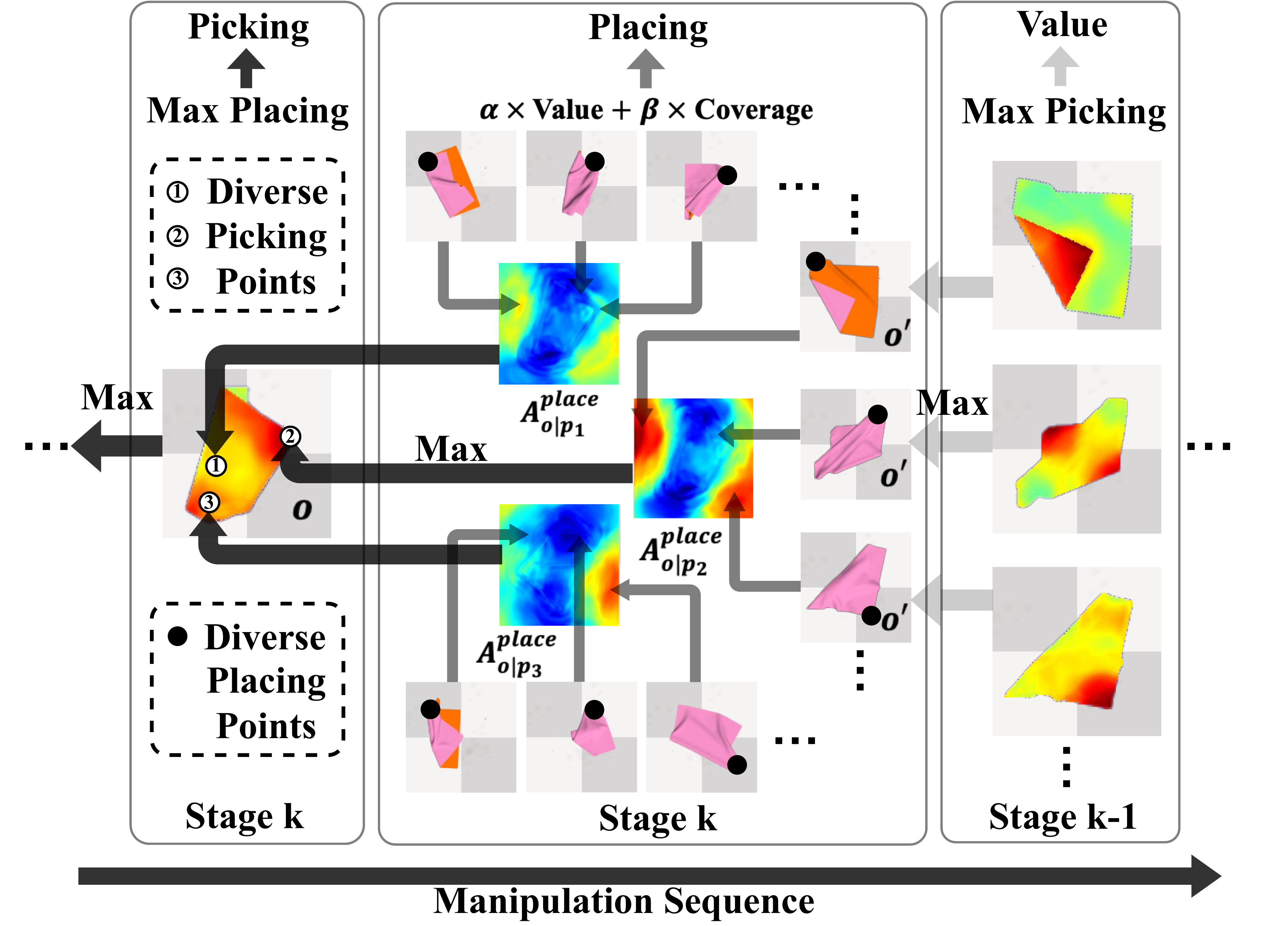}
  % \vspace{-3mm}
  \caption{\textbf{Learning placing and picking affordance with state `value's for the future.} 
  Left to Right: The bottom black arrow indicates the manipulation (inference) order. Right to Left: Arrow flows show dependencies among placing affordance, picking affordance and `value's. 
  Given observation $o$, we select 3 picking points $p_1$ $p_2$ $p_3$, and show how to supervise corresponding placing affordance $A_{o|p_1}^{place}$ $A_{o|p_2}^{place}$ $A_{o|p_3}^{place}$, and how to supervise $A_{o}^{pick}$ on $p_1$ $p_2$ $p_3$ using computed corresponding placing affordance.
  }
  \label{fig_train}
\vspace{-2mm}
\end{figure}

Like other dense affordance studies~\cite{mo2021where2act, wu2022vatmart, zhao2022dualafford}, a \textbf{greedy} way to supervise $A_{o|{p_{pick}}}^{place}$ is directly using the distance between \textbf{the target $T$} and \textbf{the new object state $o^{\prime}$} after picking $p_{pick}$ and placing on $p_{place}$ in $o$. 
For example, we use $1 - dist(o,\ T)$, \emph{i.e.}, the cloth coverage area, to supervise $A_{o|{p_{pick}}}^{place}$ for unfolding.
So we can estimate placing affordance score $g_{o,\ p_{place}|p_{pick}}^{place}$ on $p_{place}$ as (Figure~\ref{fig_train}, Middle, temporarily dismiss `value' in the Figure):
\vspace{-2mm}

\begin{equation}
\label{eq1}
g_{o,\ p_{place}|p_{pick}}^{place} = 1 - dist(o^{\prime},\ T)
\end{equation}

Given a picking point $p_{pick}$, the placing policy will select $p_{place}$ with the highest affordance, so the picking affordance score $g_{o,\ p_{pick}}^{pick}$ on $p_{pick}$
 can be estimated using the affordance score of the best placing point (Figure~\ref{fig_train}, Left):

\vspace{-4mm}

\begin{equation}
\label{eq2}
g_{o,\ p_{pick}}^{pick}=\max_i{g_{o,\ p_i|p_{pick}}^{place}}, i \in \{1, .., m\times n\}
\end{equation}

\vspace{-1mm}

We use two networks $\mathcal{M}_{pick}$ and $\mathcal{M}_{place}$ to respectively learn $A_{o}^{pick}$ and $A_{o|{p_{pick}}}^{place}$ (architectures in Section~\ref{sec:network}).

During inference, in each step, the greedy policy first selects $p_{pick}$ with the highest affordance score in $A_{o}^{pick}$ given $o$, and then selects $p_{place}$ with the highest affordance score in $A_{o|{p_{pick}}}^{place}$ given $o$ and $p_{pick}$, resulting in the \textbf{temporary best state} after the pick-and-place action.

\vspace{-1mm}
\subsection{Estimating State Values and Learning Foresightful Affordance}
\label{sec:value}
\vspace{-1mm}

As shown in Figure~\ref{fig_global_local}, for multi-step manipulation, only evaluating the direct distance between the current state and the target 
(greedy method described above) may result in many local optimal states that are temporarily closer to target but harder for future actions to complete the whole task.

Dynamic Programming (DP)~\cite{bellman1966dynamic} and Q-Learning~\cite{watkins1992q} tackle this local optima problem by estimating the \textbf{`value'} of a state that indicates whether a state is beneficial to the task in the long term (instead of the current performance).
Inspired by them,
we can add such state `value' (formally formulated in Equation~\ref{eq4}) to the estimation of $A_{o|{p_{pick}}}^{place}$:

\vspace{-3mm}
\begin{eqnarray}
\label{eq3}
g_{o,\ p_{place}|p_{pick}}^{place} = \alpha \times value_{o^{\prime}} + \beta \times (1 - dist(o^{\prime},\ T))
\\
\mbox{where}\quad \alpha + \beta = 1 \nonumber
\end{eqnarray}

With such $A_{o|{p_{pick}}}^{place}$ that is foresightful for long-term tasks,
$A_{o}^{pick}$,
which is the aggregation of $A_{o|{p_{pick}}}^{place}$,
will therefore get such foresightfulness spontaneously.

Estimating the `value' in a state requires understanding all possible actions and their corresponding future results, and then selecting the best for the estimation.
As $A_{o}^{pick}$ estimates the action result on each point,
state `value' can be estimated by selecting the $p_{pick}$ with the highest score in $A_{o}^{pick}$ (Figure~\ref{fig_train}, Left):

\begin{figure}[t]
  \centering
  \includegraphics[width=\linewidth]{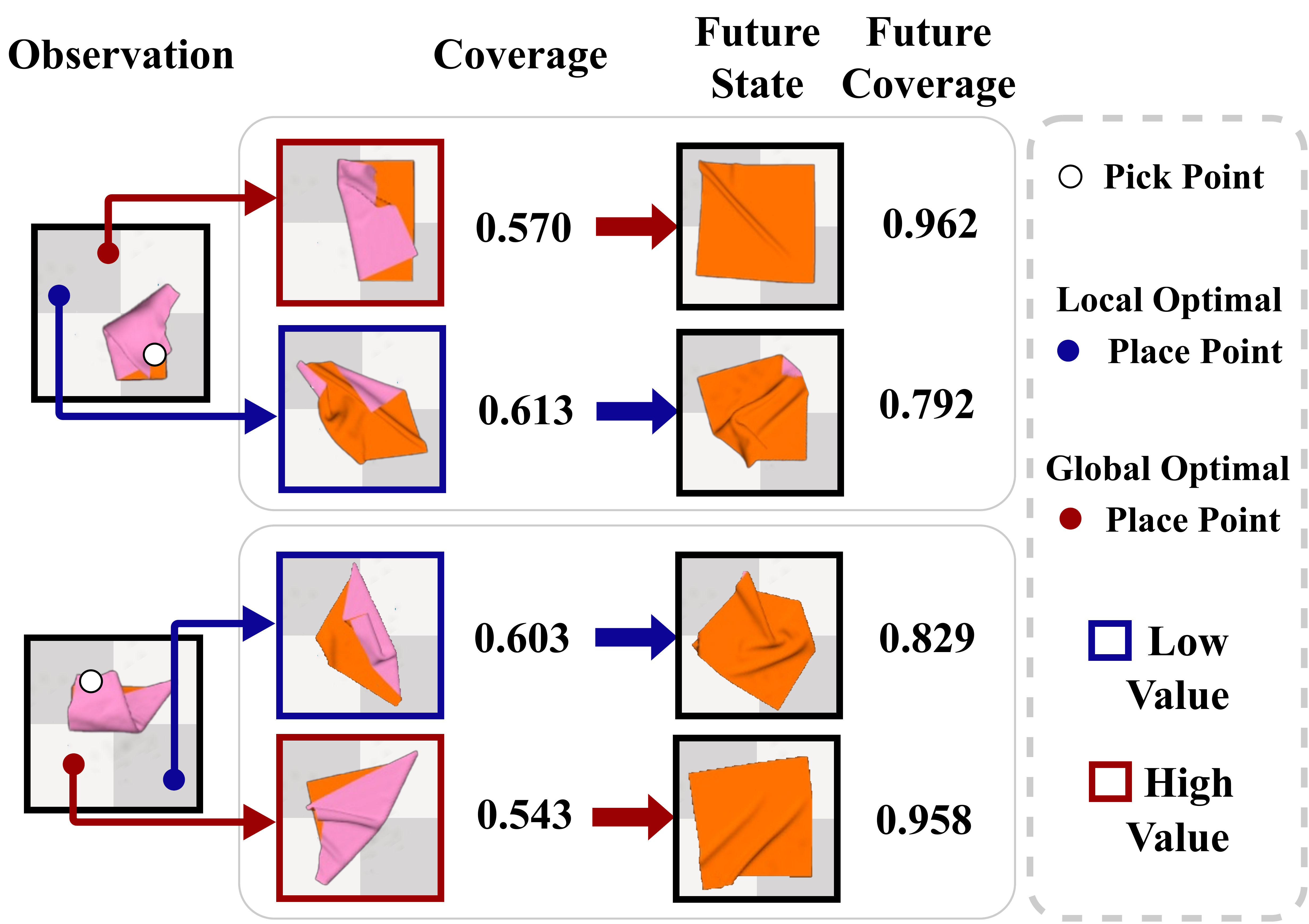}
  \caption{\textbf{Local Optima v.s. Global Optima.} Many local optimal states are temporarily closer to target (\emph{e.g.}, having larger coverage area in unfolding task), but making future actions hard to coordinate to accomplish the whole task. We propose to use `value' to indicate whether a state is suitable for future actions, with which the policy can avoid those local optimal states in multi-step tasks.
}
\vspace{-3.9mm}
  \label{fig_global_local}
\end{figure}

\vspace{-3mm}
\begin{equation}
\label{eq4}
value_{o}=\max_i{g_{o,\ p_i}^{pick}}, i \in \{1, .., m\times n\}
\end{equation}

As $value_{o}$ could be estimated by $A_{o}^{pick}$, $A_{o|{p_{pick}}}^{place}$ could be reformulated using both $A_{o}^{pick}$ and the direct distance:

\vspace{-5mm}
\begin{equation}
\label{eq5}
\begin{split}    
g_{o,\ p_{place}|p_{pick}}^{place} &= \alpha\times\max_i{g_{o^{\prime},\ p_i}^{pick}} \\
&+ \beta \times (1-dist(o^{\prime},T)) 
\\
&\mbox{where} \quad \alpha + \beta = 1
\end{split}
\end{equation}
\vspace{-5mm}

\vspace{-1mm}
\subsection{\textit{Break the Cycle and Cut into Stages}: Learning Foresightful Affordance Stably Stage by Stage}
\label{sec:learn_aff}
\vspace{-1mm}

The above-formulated picking and placing affordance forms a chicken-egg dependency cycle: 
picking affordance is dependent on placing affordance, while placing affordance is dependent on picking affordance.

To break the dependency cycle, \textbf{states close to the target} (\emph{e.g.}, cloth almost fully unfolded) become the key.
As the task is almost accomplished in these states, their values and direct distances to the target are nearly the same.
So for an interaction where the ending state $o^\prime$ is close to the target, we directly use their distances to the target (instead of both distances and `value's) to supervise the corresponding placing affordance of the starting state $o$:

\vspace{-4mm}
\begin{equation}
\label{eq6}
\begin{split}
&g_{o,\ p_{place}|p_{pick}}^{place}=1 - dist(o^{\prime},\ T)\\
&\mbox{when}\ dist(o^{\prime},\ T)\ \mbox{is close to 0}\quad i.e.\mbox{, the last step}
\end{split}
\end{equation}

According to Equations~\ref{eq2}~\ref{eq5}~\ref{eq6}, the proposed picking and placing affordance could be estimated without the dependency cycle.
As the dependency cycle breaks in states close to the target, we learn dense affordance from these simple states to more complex states (reversed order of inference) as shown in Figure~\ref{fig_framework} (Right).
Specifically,
we divide the learning procedure into multiple stages. 
In the first stage, we learn affordance for states that can reach states close to the target within one step,
using the direct distances of the following states as supervisions. 
In the $i$-th ($i>1$) stage,
we learn affordance for states that can reach states in the ($i$-1)-th stage within one step,
using both the direct distances and the `value's of states in the ($i$-1)-th stage as supervisions.
In each stage,
we first train $\mathcal{M}_{place}$ using stable `value's provided from the trained $\mathcal{M}_{pick}$ in the previous stage,
and then train $\mathcal{M}_{pick}$ with stable supervisions (max of placing affordance) provided from the trained $\mathcal{M}_{place}$ in this stage.

In this way, during the whole training, 
$A_{o|{p_{pick}}}^{place}$ and $A_{o}^{pick}$ both have stable supervisions from the former stage, and can provide stable supervisions for the latter stage.

This stage-by-stage learning schema empowers our method with superiority over RL methods. 
Although RL also estimates states and actions using Bellman Equation, 
it simultaneously estimates and updates values across all states (offline RL) or trajectories of states (online RL), 
which is difficult to efficiently and stably learn the values, 
as RL struggles in iteratively updating state `value's, especially when considering the prohibitively large state and action space of deformable objects.
In contrast,
like other dense affordance works~\cite{mo2021where2act, zhao2022dualafford, wu2022vatmart},
we stably learn affordance with `value' using supervised learning, 
with stable supervisions provided in the previous training stage.
Experiments demonstrate our superiority over RL in Section~\ref{exp_rl}.

\subsection{\textit{Fold to Unfold}: Efficient Multi-stage Data Collection for Learning Foresightful Affordance}
\label{sec:data}

The above-described training schema requires multi-stage data for training.
Specifically,
in the first stage,
the starting states are one-step to states close to the target.
In the $i$-th ($i>1$) stage,
the starting states are one-step to states in the ($i$-1)-th stage.
Each starting state's actions and corresponding ending states should be diverse, as we need to learn the affordance representing dense distributions.
% Although the ending states in each stage may be diverse, the `value's will be provided from the best of next states (shown in red squares in Figure~\ref{fig_framework}, Left).

However, due to the complexity of states and dynamics,
data collection methods used by previous dense affordance works (random policies~\cite{mo2021where2act, mo2021o2oafford} or state-based RL~\cite{wu2022vatmart, zhao2022dualafford}) could hardly collect such data.
On the other hand,
designing expert policies~\cite{seita2021learning} or hand-crafting demonstrations are difficult and time-consuming for different tasks.

Therefore, we propose a novel self-supervised method, named \textit{Fold to Unfold}, using reversed actions of tasks to efficiently collect multi-stage data.
This method is generic to many kinds of deformable object manipulation tasks, with no need for human-designed expert policies or annotations.

\begin{figure}[h]
  \centering
  \includegraphics[width=\linewidth]{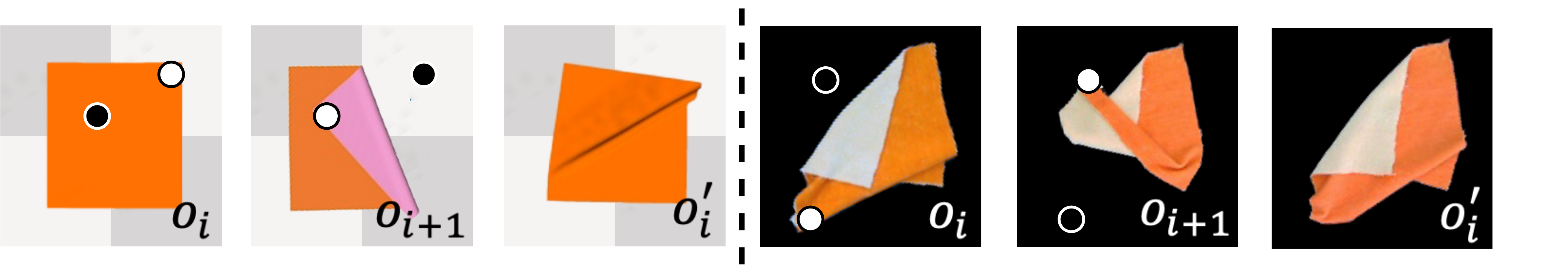}
  \caption{\textit{Fold to Unfold} collection in simulator and real world.}
\vspace{-3mm}
  \label{fig_fold_unfold}
\end{figure}

Similar to the training procedure,
we collect data from states close to the target, to more complex states.

Specifically, as shown in Figure~\ref{fig_fold_unfold}, from a state $o_i$ in stage $i$, we select a picking point $p_{pick}$, put it on a placing point $p_{place}$ and get $o_{i+1}$. 
Then, from $o_{i+1}$, we execute the reversed action, pick $p_{place}$, place on $p_{pick}$ and get $o_i^{\prime}$.
If $o_i^{\prime}$ is similar to $o_i$, we choose $o_{i+1}$ as a starting state in stage $i$+1, and sample diverse actions on $o_{i+1}$ with different corresponding results to train dense affordance in the ($i$+1)-th stage (shown in Figure~\ref{fig_framework}).

Through a few stages of data collection, object states become complex and diverse,
empowering trained affordance networks with generalization towards diverse novel states.

\textbf{Note that}, although reversed actions cannot fully recover previous states, \emph{i.e.}, $o_i^{\prime}$ are not the same as $o_i$, chances are that $o_i^{\prime}$ and $o_i$ are similar, and thus this method still greatly improves sample efficiency.
Also, proposed affordance is \textbf{not dependent on} this data collection method, as it can be trained on data collected by any method.

\vspace{-1mm}
\subsection{Integrated Systematic Training}
\label{sec:ist}
\vspace{-1mm}

Although the above designs and training procedure enable the affordance learning for multi-step tasks, $\mathcal{M}_{pick}$ and $\mathcal{M}_{place}$ are trained using only offline collected data in different stages, without considering the actual execution performance of the policies provided by the two modules.
During actual manipulation procedures, the policy is the composite policy of $\mathcal{M}_{pick}$ and $\mathcal{M}_{place}$, and pick-and-place actions are executed one by one sequentially as a whole system.
Therefore, we propose the Integrated Systematic Training (IST) procedure to adapt $\mathcal{M}_{pick}$ and $\mathcal{M}_{place}$ using online data.

In this procedure, with offline trained $\mathcal{M}_{place}$ and $\mathcal{M}_{pick}$, 
we randomly sample object initial states,
use $\mathcal{M}_{place}$ and $\mathcal{M}_{pick}$ as the policy to select $p_{pick}$ and $p_{place}$, execute pick-and-place step by step, and use actual results to simultaneously update $\mathcal{M}_{place}$ and $\mathcal{M}_{pick}$.
Through this procedure, the two modules are constantly adapted by consecutively online-sampled and actually-executed data, and thus are gradually integrated into a whole system.

\vspace{-1mm}
\subsection{Network Architectures and Loss Function}
\label{sec:network}
\vspace{-1mm}

For architectures of $\mathcal{M}_{pick}$ and $\mathcal{M}_{place}$, 
we use Fully Convolutional Networks (FCNs)~\cite{long2015fully} same in Transporter~\cite{zeng2021transporter, seita2021learning} with extra skip-connections as backbone per-point feature extractor. 
For $\mathcal{M}_{pick}$, we directly use $p_{pick}$ feature to predict picking affordance score on $p_{pick}$.
For $\mathcal{M}_{place}$, we use feature concatenation of $p_{pick}$ and $p_{place}$ to predict placing affordance score on $p_{place}$.
To train both $\mathcal{M}_{pick}$ and $\mathcal{M}_{place}$, we use Mean Absolute Error (MAE) between predictions and ground truth as the loss function.

% \vspace{-3mm}
\section{Experiments}
% \vspace{-2mm}

\subsection{Tasks, Settings and Metrics}

\paragraph{Tasks.} To demonstrate the superiority of our framework, we select 2 representative tasks from \textit{DeformableRavens} benchmark: (1) \textbf{cable-ring}: manipulating a ring to a given green circle, (2) \textbf{cable-ring-notarget}: manipulating a ring to any circle, as well as 2 representative tasks from \textit{SoftGym}: (3) \textbf{SpreadCloth}: spreading crumpled cloth to be flat, (4) \textbf{RopeConfiguration}: manipulating a rope from a random pose to a target pose (we use the shape `S' as the target). Among them, the first two are relatively easier. They can be accomplished without considering future actions and states, and we conduct them to show our dense affordance's superiority over methods imitating expert demonstrations. The last two are much harder and would be better accomplished considering future actions and states to avoid local optima.

\vspace{-5mm}

\paragraph{Settings.} For all tasks, in both training and testing, we set different random seeds for each episode, producing unseen and diverse initial poses of objects. To compare the generalization ability between our proposed dense affordance and imitation-based methods, for cables in \textit{DeformableRavens}, we directly test the model trained on cables with 32 beads over novel cable configurations with 24, 28 or 36 beads.

\vspace{-5mm}

\paragraph{Metrics.} For cable-ring and cable-ring-notarget, we follow \textit{DeformableRavens} and use the manipulation successful rate as the metric. For SpreadCloth and RopeConfiguration, we follow \textit{SoftGym} and use the normalized score as the metric. For all tasks, higher scores indicate better performance.

% \vspace{-1mm}
\subsection{Baselines}
% \vspace{-1mm}

For two cable-ring related tasks, we compare our method with baselines with or without expert demonstrations:
% with and without human expert demonstrations:
%
\begin{itemize}
  \vspace{-2mm}
  \item 
  \textbf{Transporter}~\cite{zeng2021transporter, seita2021learning} is commonly used for robotic manipulation by learning visual correlation for picking and placing points. In \textit{DeformableRavens}~\cite{seita2021learning} it is trained by cloning expert demonstrations and achieves SOTA performance over relevant tasks.
  \vspace{-1mm}
  \item 
  \textbf{GT-State} receives ground truth (GT) pose of the target object, and regresses $p_{pick}$ and $p_{place}$ with MLP.
  \vspace{-1mm}
  \item 
  \textbf{GT-State 2-Step} first regresses $p_{pick}$ and then $p_{place}$ using $p_{pick}$ and GT pose concatenation, both via MLP.
\end{itemize}

For SpreadCloth and RopeConfiguration, as object states and dynamics are too complex and the tasks are too difficult to hand-engineer expert policies, we compare our method with baselines focused on multi-step planning:

\begin{itemize}
  \vspace{-2mm}
  \item 
  \textbf{CURL-SAC}~\cite{laskin_srinivas2020curl} that uses a model-free RL approach with contrastive unsupervised representations.
  \vspace{-2mm}
  \item 
  \textbf{DrQ}\cite{yarats2021drq} applies augmentation, regularization to RL.
  \vspace{-2mm}
  \item 
  \textbf{PlaNet}~\cite{hafner2019planet} learns state space dynamics for planning.
  \vspace{-2mm}
  \item 
  \textbf{MVP}~\cite{wu2019learning} learns pick-and-place policy with model-free RL designed for deformbale object manipulation.
\end{itemize}

\vspace{-0.1cm}
\subsection{Qualitative Results and Analysis}
\vspace{-0.1cm}

\begin{figure}[h!]
  \centering
  \includegraphics[width=\linewidth, trim={0.cm, 0cm, 0cm, 0cm}, clip]{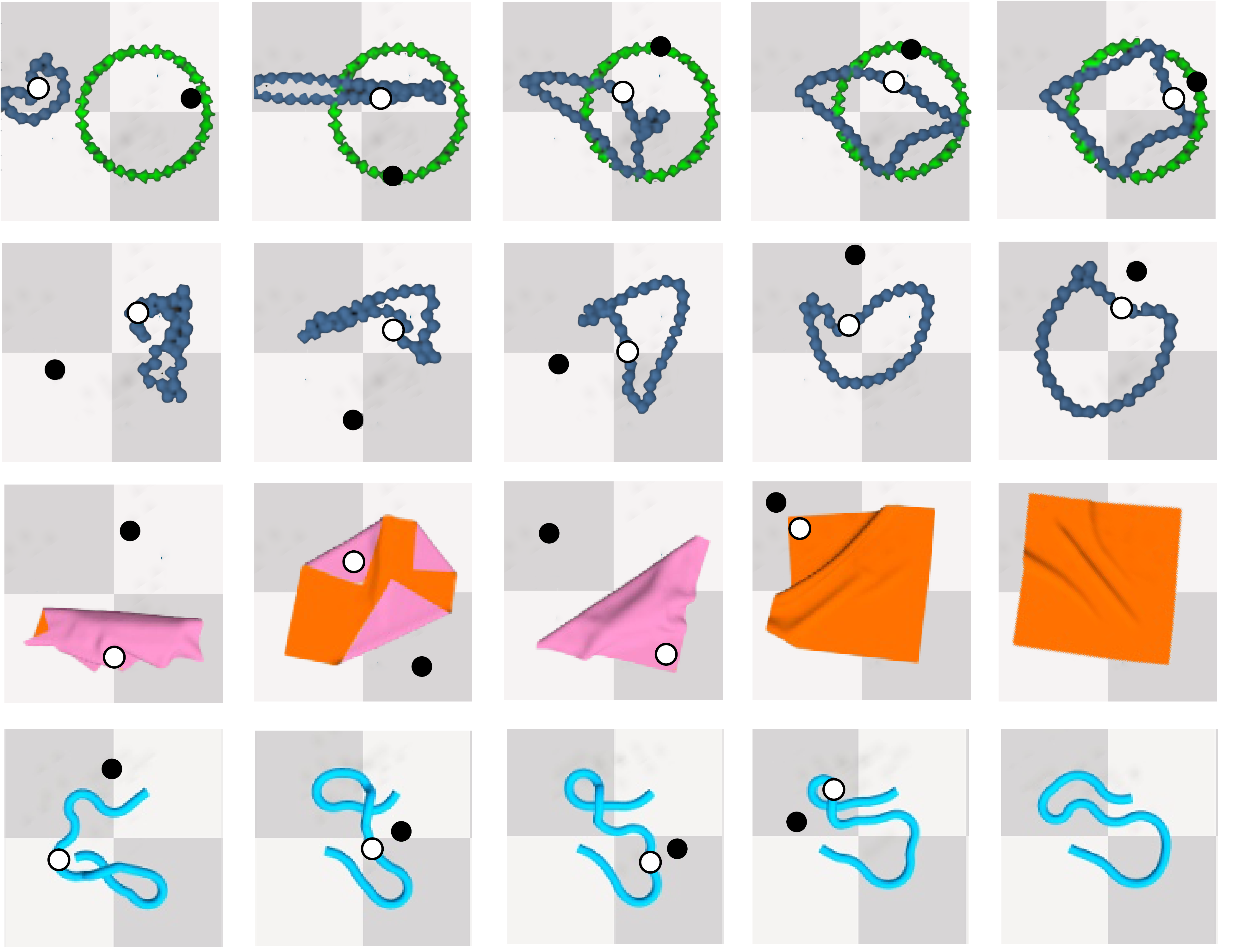}  
   \caption{\textbf{Example action sequences} for cable-ring, cable-ring-notarget, SpreadCloth and RopeConfiguration. \textbf{White point denotes picking and black point denotes placing}.}
  \label{fig_vis_traj}
  \vspace{-2mm}
\end{figure}

Figure~\ref{fig_vis_traj} shows examples of manipulation trajectories for diverse tasks using our proposed affordance.
It is worth mentioning that, in the second state of SpreadCloth (Row 3), though it is intuitive to place the picking point (white) to the top-left position, the model places it to the bottom-right position (black), as the corresponding next state has \textbf{low coverage} but \textbf{high `value'}, requiring only one following pick-and-place action to almost fully unfold the cloth.

% trim={<left> <lower> <right> <upper>}
\begin{figure}[h!]
  \centering
  \includegraphics[width=\linewidth, trim={2cm, 0cm, 0cm, 0cm}, clip]{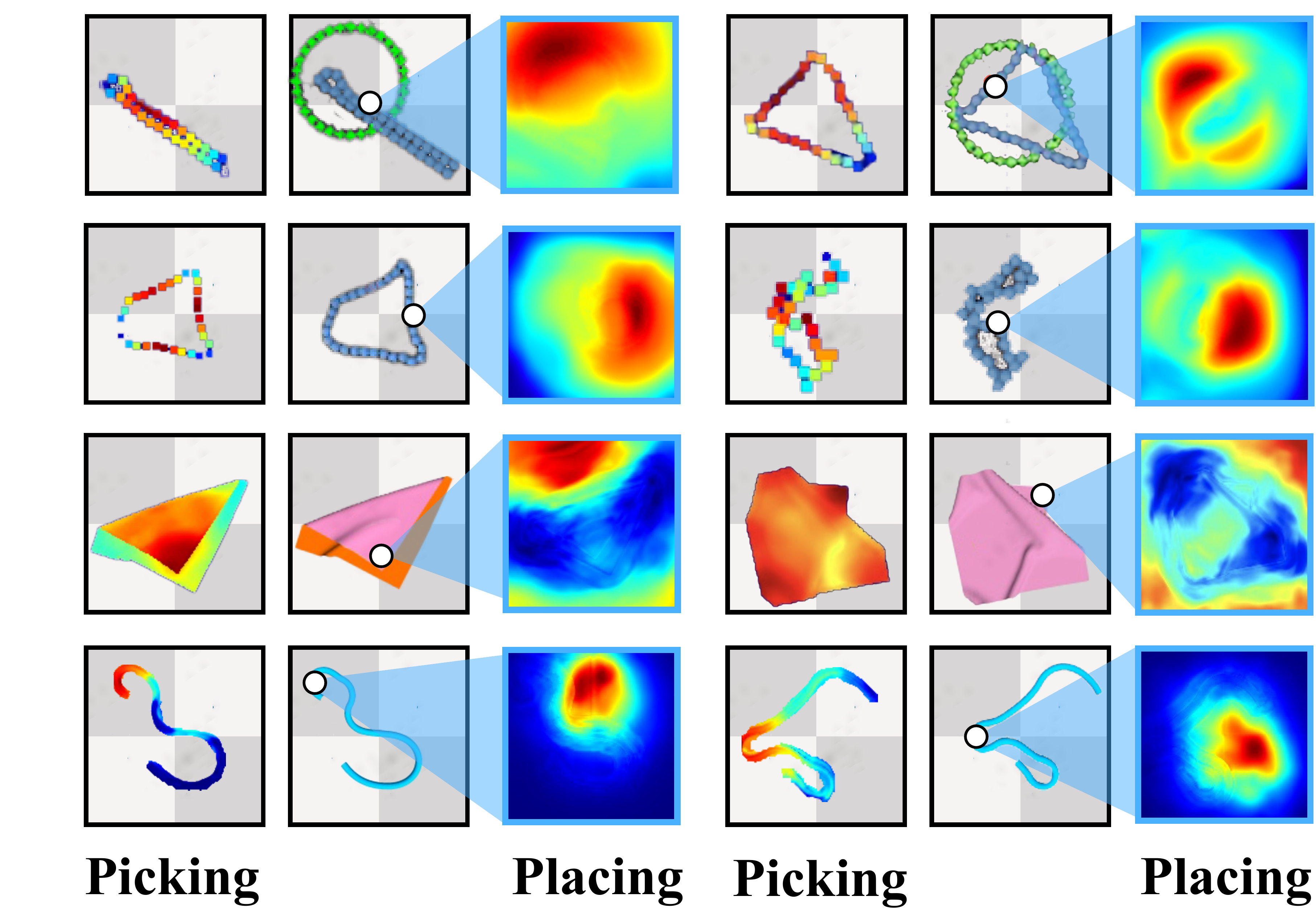}  
  \caption{\textbf{Picking and placing affordance.} Each row contains two (picking affordance, observation with $p_{pick}$, placing affordance) tuples for a task. $p_{pick}$ is selected by picking affordance. \textbf{Higher color temperature} means \textbf{higher affordance}. 
  }
  \label{fig_vis_critic}
\end{figure}

\begin{figure}[h]
  \centering
  \includegraphics[width=\linewidth, trim={0cm, 0cm, 0cm, 0cm}, clip]{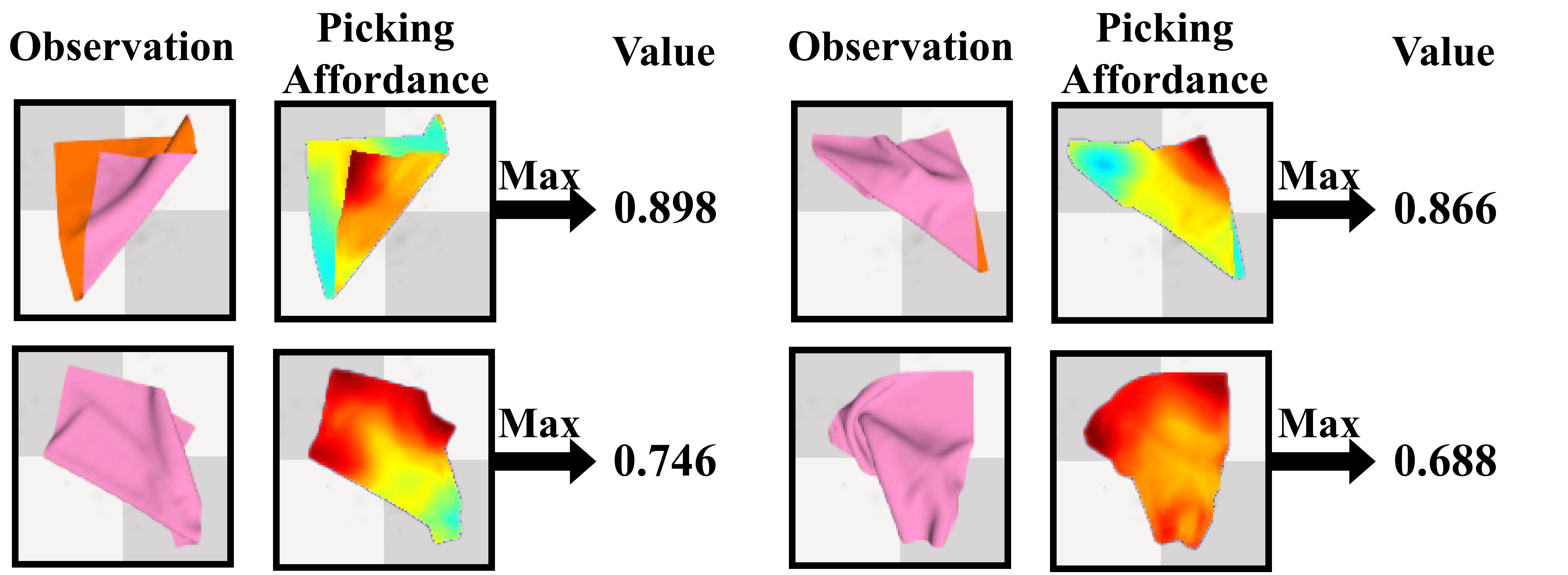}  
  \caption{\textbf{Visualization of `value'} shows that some states with closer distances to the target (\emph{e.g.}, larger area) may not have higher `value', as these states are hard for future actions to fulfill the task.
  }
  \label{fig_vis_potential}
\end{figure}

Figure~\ref{fig_vis_critic} visualizes picking and placing affordance, clearly showing that the learned affordance represents deformable objects with complex states and dynamics and facilitates selecting picking and placing points for manipulation.
Figure~\ref{fig_vis_potential} visualizes `value's of states.

% trim={<left> <lower> <right> <upper>}

\vspace{-0.1cm}
\subsection{Quantitative Results and Analysis}
\vspace{-0.1cm}
\label{exp_rl}

\begin{table}[h!]
  \centering
  \setlength{\tabcolsep}{0.9mm}
    \begin{tabular}{@{}lccccc@{}}
    \toprule
    Method              & cable-ring & cable-ring-notarget  \\ \midrule \midrule
    GT-State        &     0.0       &           5.0            \\ \midrule
    GT-State 2-Step   &      0.0      &  1.7   \\ \midrule
    Transporter      &    68.3    &   70.0      \\ \midrule
    Ours      &  \textbf{81.7}   &  \textbf{95.0}    \\ \bottomrule
    \end{tabular}
  \caption{\textbf{Quantitative results in \textit{DeformableRavens}}.}
  \label{tab_accu_single}
  \vspace{-2mm}
\end{table}

\begin{table}[h!]
  \centering
  \setlength{\tabcolsep}{0.9mm}
    \begin{tabular}{@{}lcccc@{}}
    \toprule
    Method              & SpreadCloth & RopeConfiguration \\ \midrule \midrule
    CURL-SAC     &  0.195         &      0.348  \\ \midrule
    PlaNet      &  0.387         &      0.236  \\ \midrule
    DrQ      &    0.275          &      0.154  \\ \midrule
    MVP      &    0.372          &      0.258  \\ \midrule
    Ours      &  \textbf{0.758}   &   \textbf{0.529} \\ \bottomrule
    \end{tabular}
  \caption{\textbf{Quantitative results in \textit{SoftGym}}.}
  \label{tab_accu_mul}
  \vspace{-2mm}
\end{table}

Table~\ref{tab_accu_single} shows our framework outperforms all baselines in \textit{DeformableRavens}.
Specifically, for \textbf{GT State} and \textbf{GT State 2-Step}, GT states can only provide part of the necessary information, and it is difficult to acquire the precise GT states of deformable objects in the real world.
We outperform \textbf{Transporter} that learns visual correlation for the matching between picking and placing points, even though it directly clones successful demonstrations from hand-crafted expert policies. Two possible reasons are that: 1) Dense affordance is more suitable for deformable objects as it represents the results of diverse actions on complex states, while visual correlation in \textbf{Transporter} is suitable for matching like assembling-kits;
2) training on expert demonstrations and cloning expert policies will limit the model's generalization toward diverse situations during inference.
To further evaluate the \textbf{generalization ability of the dense affordance}, we produce different novel object configurations, using 24, 28, and 36 as the bead number instead of the initial 32.
Our method's slighter decrease in performance over novel object configurations shown in Table~\ref{tab_generalization} also demonstrates its generalization ability.
Besides, expert policies need to be elaborately hand-designed for different tasks, while our method can apply to any task without modifications.

Table~\ref{tab_accu_mul} shows our framework outperforms all baselines in \textit{SoftGym}. As described in ~\ref{sec:learn_aff}, compared with those RL methods, our framework learns representations of complex states for multi-step manipulation in a stable way.

\begin{table}[h!]
  \setlength{\tabcolsep}{0.9mm}
  \centering
    \begin{tabular}{@{}lccc@{}}
    \toprule
    Task       & cable-ring & cable-ring-notarget \\
    configurations   & 24 / 28 / 36 & 24 / 28 / 36 \\ \midrule \midrule
    Transporter       &      33.3 / 58.3 / 32.7      & 60.0 / 71.7 / 31.7 \\ \midrule
    Ours        &  \textbf{61.6} / \textbf{86.7} / \textbf{58.3}   &   \textbf{81.7} / \textbf{96.7} / \textbf{78.3}  \\  \bottomrule
    \end{tabular}
  \caption{\textbf{Manipulation scores on novel configurations in \textit{DeformableRaves}} showing our method's  generalization capability.}
  \label{tab_generalization}
\end{table}

\vspace{-0.1cm}
\subsection{Ablation Studies and Analysis}
\label{sec:ablation}
\vspace{-0.1cm}

To demonstrate necessities of our framework's different components, we conduct ablation experiments by comparing our method with:
    % \item 
    1) \textbf{Ours RandPick}: our method with the picking policy replaced by a random policy;
    % \item 
    2) \textbf{Ours ExpertPick}: our method with the picking policy replaced by Transporter's expert;
    % \item 
    3) \textbf{Ours w/o IST}: our method without Integrated Systematic Training (IST);

For SpreadCloth and RopeConfiguration that require strongly related sequential actions, we additionally compare 1) ablated versions using different stages of data, 2) \textbf{Ours only dist} directly and \textbf{greedily} trained on all collected data instead of stage-by-stage considering `value's.

\begin{table}[h!]
  \centering
    \begin{tabular}{@{}lccccc@{}}
    \toprule
    Method              & cable-ring & cable-ring-notarget  \\ \midrule \midrule
    Ours RandPick        &      11.7      &          58.3                 \\ \midrule
    Ours ExpertPick        &     76.7       &        41.7                     \\ \midrule
    Ours w/o IST         &     78.3       &     91.7                  \\ \midrule
    Ours      &  \textbf{81.7}  &  \textbf{95.0}        \\ \bottomrule
    \end{tabular}
  \caption{\textbf{Ablation studies} in \textit{DeformableRavens}.}
  \label{tab_ablation}
    \vspace{-3mm}
\end{table}

\begin{table}[h!]
  \centering
    \setlength{\tabcolsep}{1.2mm}
    \begin{tabular}{@{}lccccc@{}}
    \toprule
    Method              & stage1 & stage2 & stage3 & stage4 & stage5  \\ \midrule \midrule
    Ours RandPick        & 0.241 & 0.211 & 0.304 & 0.185 & 0.190             \\ \midrule
    Ours w/o IST         & 0.526 & 0.586 & 0.621 & 0.624 & 0.612                \\ \midrule
    Ours only dist & \textbf{0.701} & \textbf{0.701} & 0.701 & 0.701 & 0.701  \\ \midrule
    Ours      &  0.589 & 0.695 & \textbf{0.752} & \textbf{0.754} & \textbf{0.758}    \\ \bottomrule
    \end{tabular}
  \caption{\textbf{Ablation studies} in \textit{SpreadCloth}.}
  \label{tab_multi_step_cloth}
    \vspace{-3mm}
\end{table}

\begin{table}[h!]
  \centering
    \setlength{\tabcolsep}{1.2mm}
    \begin{tabular}{@{}lccccc@{}}
    \toprule
    Method              & stage1 & stage2 & stage3 & stage4 & stage5  \\ \midrule \midrule
    Ours RandPick        & 0.329 & 0.302 & 0.332 & 0.334 & 0.322             \\ \midrule
    Ours w/o IST         & 0.359 & 0.418 & 0.437 & 0.479 & 0.474                \\ \midrule
    Ours only dist & \textbf{0.460} & 0.460 & 0.460 & 0.460 & 0.460  \\ \midrule
    Ours      &  0.441 & \textbf{0.503} & \textbf{0.518} & \textbf{0.527} & \textbf{0.529}    \\ \bottomrule
    \end{tabular}
  \caption{\textbf{Ablation studies} in \textit{RopeConfiguration}.}
  \label{tab_multi_step_rope}
    \vspace{-5mm}
\end{table}

Table~\ref{tab_ablation},~\ref{tab_multi_step_cloth}and~\ref{tab_multi_step_rope} show quantitative results of ablation experiments.
\textbf{Ours RandPick} and \textbf{Ours ExpertPick} show that, with the same placing affordance, our proposed picking affordance helps the framework perform the best.

\begin{figure}[h!]
  \centering 
%   \fbox{\rule[-.5cm]{0cm}{4cm} \rule[-.5cm]{4cm}{0cm}} 
 \includegraphics[width=\linewidth, trim={0cm, 0cm, 0cm, 0cm}, clip]{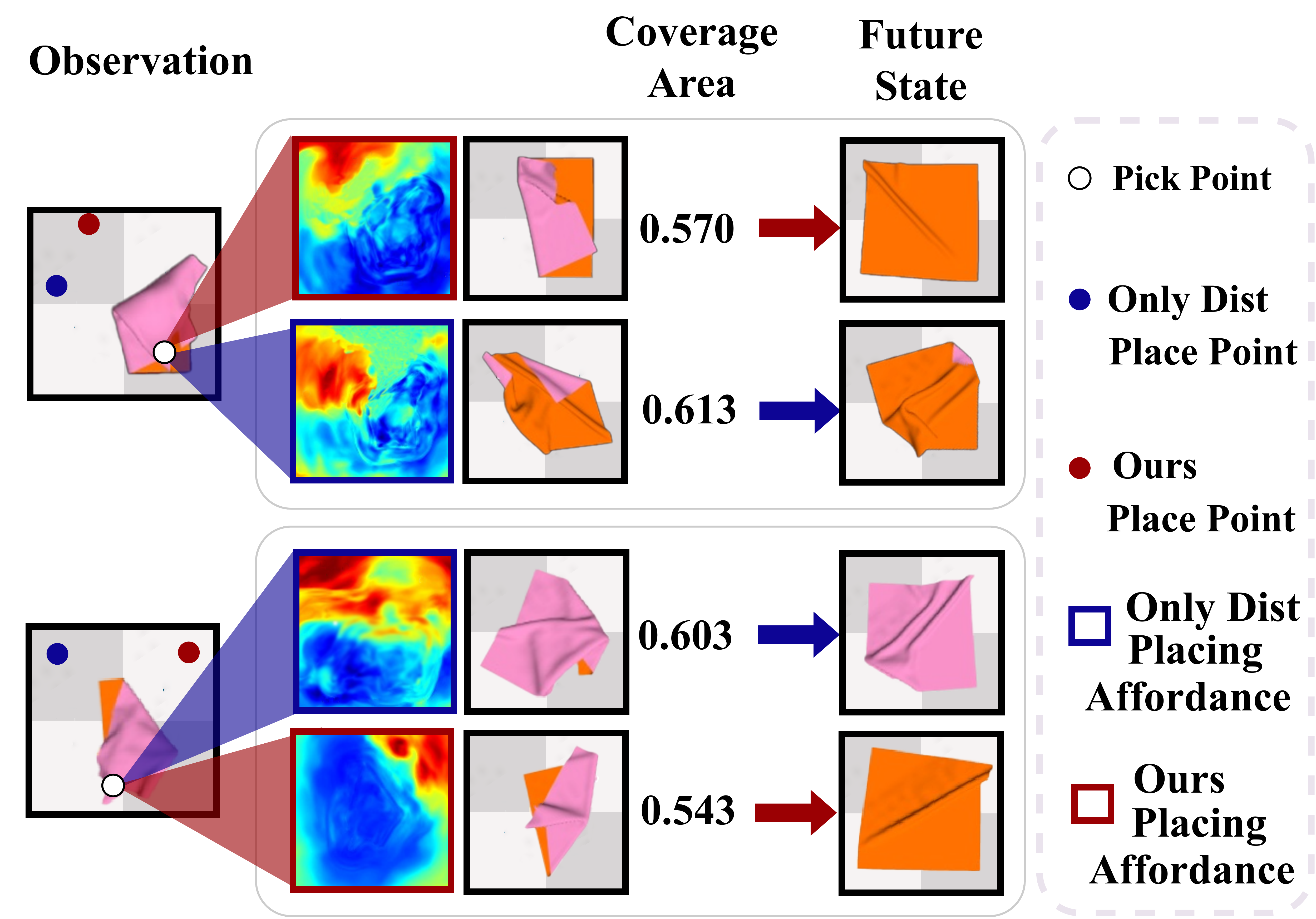}  
  \caption{\textbf{Placing affordance trained using `value' supervision ({\color{red}red}) and only using the greedy direct distance ({\color{blue}blue})}.}
   \label{fig_multi_step_vs_gt}
     \vspace{-5mm}
\end{figure}

\textbf{Ours only dist} in Table~\ref{tab_multi_step_cloth}and~\ref{tab_multi_step_rope} show that, directly training on all the diverse data without estimating state `value' limits the performance compared with our proposed framework. Besides, as shown in Figure~\ref{fig_multi_step_vs_gt}, \textbf{Ours only dist} will propose actions leading to local optimal states, while our foresightful affordance will help to avoid that.

Table~\ref{tab_multi_step_cloth} and~\ref{tab_multi_step_rope} show that, a series of steps of data empowers affordance with generalization to diverse states.

\begin{figure}[h!]
  \centering 
%   \fbox{\rule[-.5cm]{0cm}{4cm} \rule[-.5cm]{4cm}{0cm}}
 \includegraphics[width=\linewidth, trim={0cm, 0mm, 0cm, 0cm}, clip]{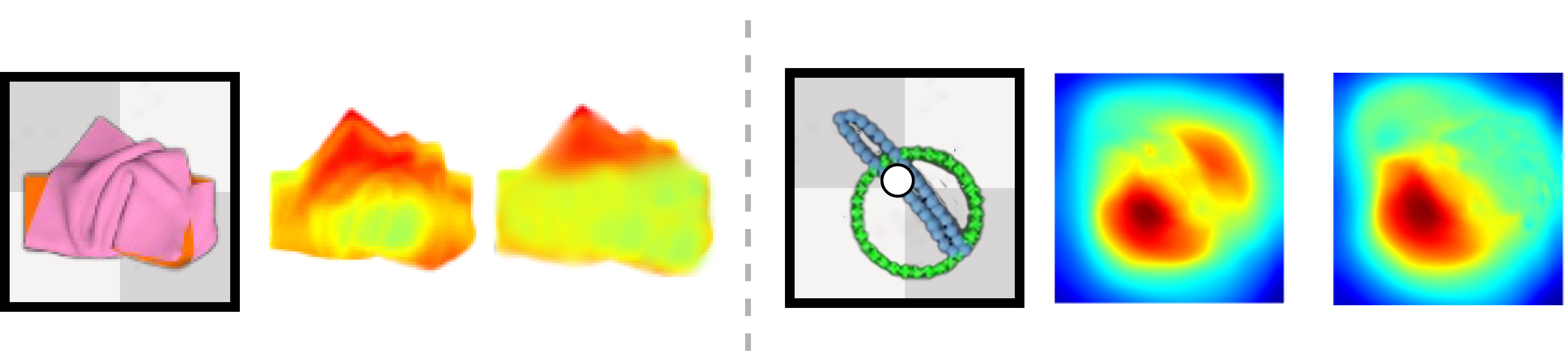}  
  \caption{\textbf{Picking and placing affordance before (middle) and after (right) IST} of the observation (left). White: pick point.}
   \label{fig_joint_training}
   \vspace{-3mm}
\end{figure}

\textbf{Ours w/o IST} in Table~\ref{tab_ablation},~\ref{tab_multi_step_cloth}and~\ref{tab_multi_step_rope}, and adjusted affordance and `value's in Figure~\ref{fig_joint_training} demonstrate IST helps integrating picking and placing modules into a whole, generating more precise perception of affordance and `value's.

% trim={<left> <lower> <right> <upper>}

% trim={<left> <lower> <right> <upper>}

\subsection{Real-world Experiments}

To bridge the sim2real gap and implement our method in the real world, similar to~\cite{wu2019learning, ha2021flingbot}, we use domain randomization to train affordance models in simulation and fine-tune them in real world.
Specifically, we collect real-world data using \textit{Fold to Unfold}, fine-tune trained-in-simulation $\mathcal{M}_{pick}$ and $\mathcal{M}_{place}$ stage by stage using the collected data.

For evaluations, we randomly lift and drop the objects for five times to get the initial state, and then run the models for ten pick-and-place actions to perform the tasks. The manipulation score is the average normalized score (computed the same as in \textit{SoftGym}) of sixty trajectories. 

\begin{figure}[h!]
  \centering 
%   \fbox{\rule[-.5cm]{0cm}{4cm} \rule[-.5cm]{4cm}{0cm}} 
 \includegraphics[width=\linewidth, trim={0cm, 0cm, 0cm, 0cm}, clip]{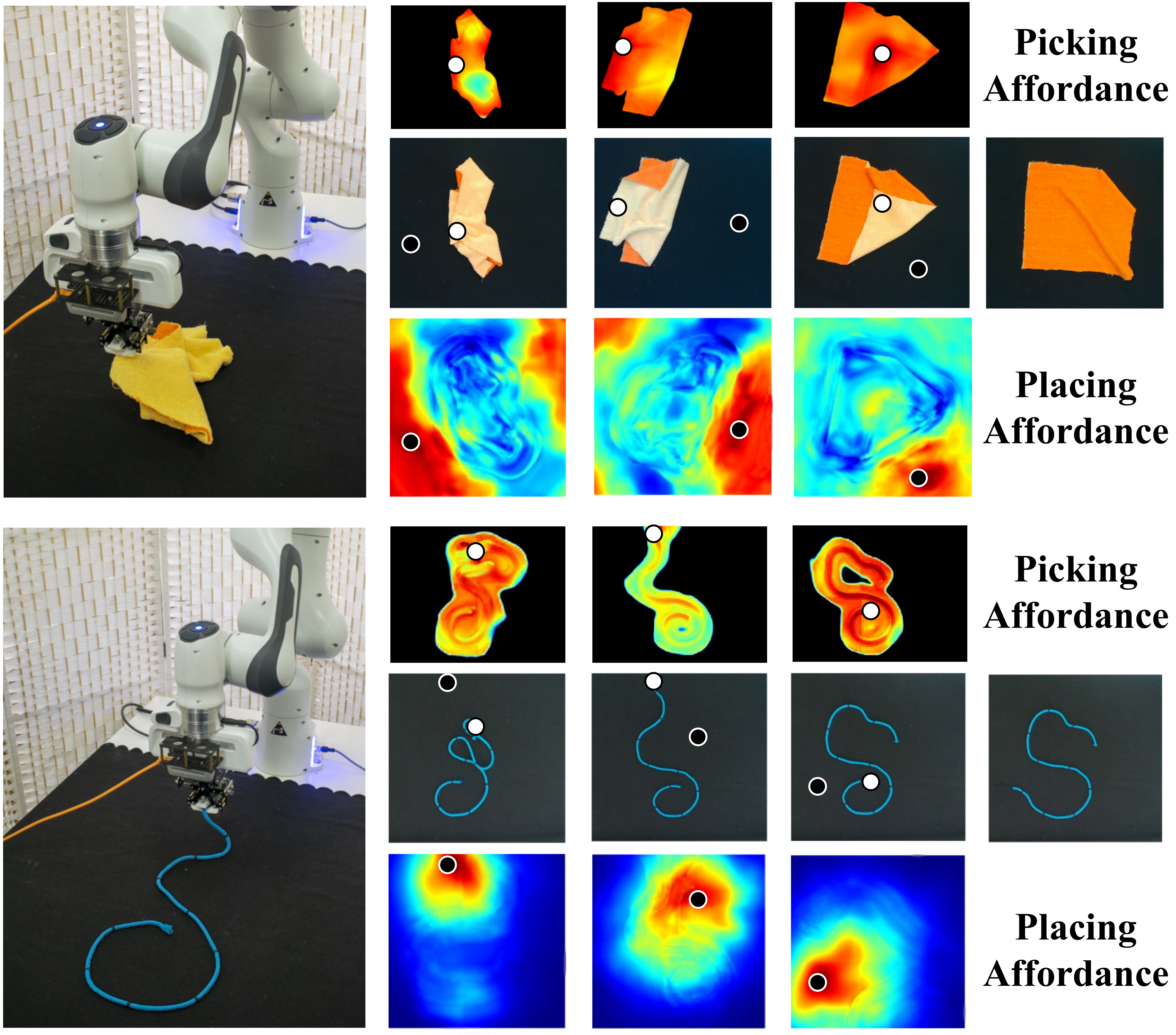}  
  \caption{Examples of real-world manipulation trajectories guided by picking and placing affordance.}
   \label{fig_real_world}
     \vspace{-2mm}
\end{figure}

Shown in Figure~\ref{fig_real_world}, given real-world observations with textures and physics different from objects in simulation, our method predicts reasonable picking and placing affordance and selects actions for tasks.

We compare our method with MVP~\cite{wu2019learning}, as it also uses pick-and-place as action primitive and thus could do both cloth and rope related tasks, and provides real world experiments. 
As shown in Table~\ref{tab_real_world}, our method outperforms MVP as explained in ~\ref{exp_rl}.

See the supplementary for more implementation details.

\begin{table}[h!]
  \centering
    \begin{tabular}{@{}lccccc@{}}
    \toprule
    Method              & SpreadCloth  & RopeConfiguration \\ \midrule \midrule
    MVP        &      0.307      &         0.227       \\ \midrule
    Ours        &     \textbf{0.683}       &        \textbf{0.461}         \\ \midrule
    \end{tabular}
  \caption{Manipulation scores in the real world.}
  \label{tab_real_world}
    \vspace{-3mm}
\end{table}

% trim={<left> <lower> <right> <upper>}

\vspace{-3mm}
\section{Conclusion}
\label{sec:conlusion}
\vspace{-1mm}

We propose to use dense visual affordance for manipulating deformable objects with complex states and dynamics.
For tasks that require a series of strongly related actions, we further empower the proposed affordance with the awareness of a certain action's influence on subsequent actions.
We propose a self-supervised framework with novel designs to efficiently collect multi-stage interaction data and stably learn this representation. Experiments on representative tasks and in the real world show the superiority of our proposed dense affordance and the learning framework.

{\small
\bibliographystyle{ieee_fullname}
\bibliography{reference}
}

% \clearpage

\appendix

\section{Additional Details of Tasks, Settings and Metrics}

\subsection{Tasks}
We select 2 representative tasks from \textit{DeformableRavens} benchmark: \textbf{cable-ring} and \textbf{cable-ring-notarget}, as well as 2 harder tasks from \textit{SoftGym}: \textbf{SpreadCloth} and \textbf{RopeConfiguration} (we use the shape `S' as the target). 
% The aim of the first two tasks is to show our proposed framework can learn meaningful and abundant information without human-designed policy to . We compare our method with imitation learning in these two tasks, which shows our framework 

\begin{itemize}
 \item 

 (1) For \textbf{cable-ring}, it has a ring-shaped cable with 32 beads. The goal of the robot is to manipulate the cable towards a target zone denoted by a green circular ring in the observation. The maximum convex hull area of the cable-ring and the target cable-ring are the same.
% The termination condition is based on whether the area of the convex hull of the cable
% is above a threshold.
 \item 
(2) For \textbf{cable-ring-notarget}, the setting is the same as \textbf{cable-ring}, except that there is no visible target zone in the observation, so that the goal is to manipulate the cable to a circular ring anywhere on the workbench.
 \item 
(3) For \textbf{SpreadCloth}, we use a square cloth. The cloth is randomly perturbed to a crumpled state. The goal of the robot is to manipulate the crumpled cloth into flat state.
 \item 
(4) For \textbf{RopeConfiguration}, the rope is randomly perturbed to a crumpled state. The goal of the robot is to manipulate the rope into the shape of letter 
`S'.
\end{itemize}

\begin{figure}[htbp]
  \centering
  \includegraphics[width=\linewidth, trim={0cm, 12cm, 10cm, 0cm}, clip]{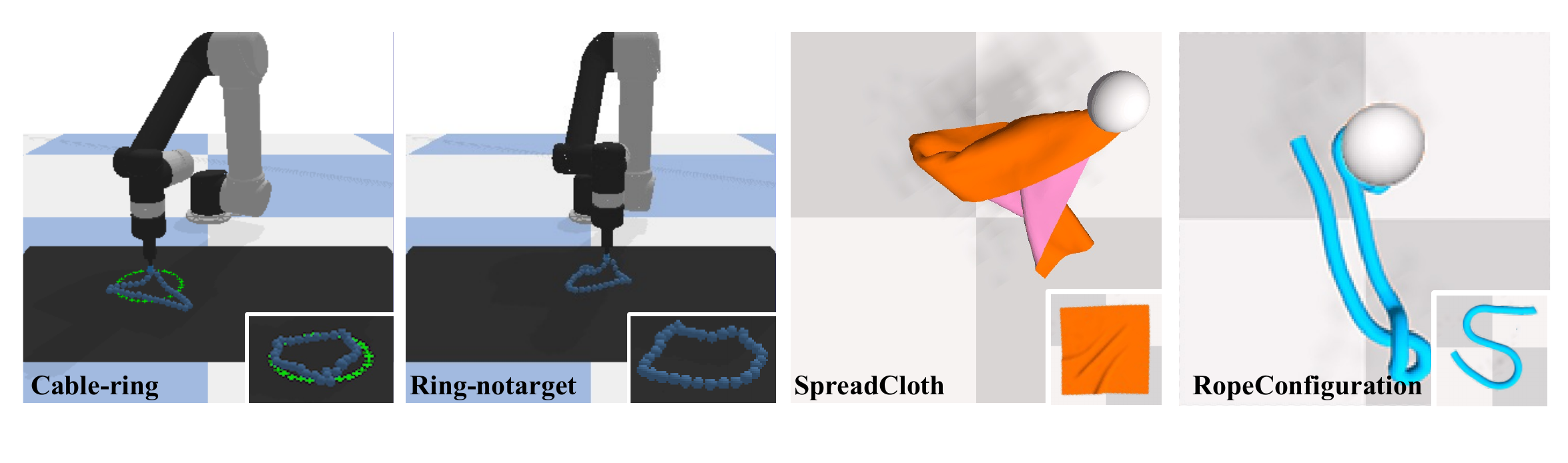}
  \caption{\textbf{Demonstrations of the selected tasks.} Each image shows an observation and a successful state (the cropped sub-images to the bottom right).}
  \label{fig:tasks}
\end{figure}

\subsection{Simulation}
For \textit{DeformableRavens} benchmark, we use the suite of simulated manipulation tasks using PyBullet~\cite{coumans2016pybullet} physics engine and OpenAI GYM~\cite{1606.01540} interfaces. For \textit{SoftGym} benchmark, we use Nvidia FleX physics simulator and the Python interface for a better simulation of cloth and rope.

\subsection{Metrics}
\label{sec:simu_metric}
For \textbf{cable-ring} and \textbf{cable-ring-notarget}, we follow \textit{DeformableRavens} benchmark, when the convex hull area of the ring beads exceeds a thresh $\beta$, the manipulation is judged as a success. We set $\beta$ to be 0.75 for these two tasks, as established in \textit{DeformableRavens}. We use the success rate as manipulation score, which is number of successful manipulation trajectories divided by total number of manipulation trajectories. 

For the \textbf{SpreadCloth} and \textbf{RopeConfiguration}, following \textit{SoftGym} benchmark, we choose the normalized score as manipulation score, which is $\frac{metric_{final}-metric_{initial}}{metric_{goal}-metric_{initial}}$, where $metric_{final}$ means the score of final state, $metric_{initial}$ means the score of initial state and $metric_{goal}$ means the score of target state. Specifically, for \textbf{SpreadCloth}, we use the coverage area of cloth as the measurement, and $metric_{goal}$ is $1.00$. For \textbf{RopeConfiguration}, we use the negative bipartite graph matching distance as the metric, and $metric_{goal}$ is $-0.04$.

During \textbf{testing}, we randomly select 60 (the same number of trials as in \textit{DeformableRavens}) random seeds representing different initial configurations of the objects, conduct experiments and report manipulation score on these different initial states.

\section{Additional Details of Experiments}

\subsection{Data Collection}
We collect 5,000 interactions in cable-related tasks, and 40000 interactions in \textbf{SpreadCloth} and \textbf{RopeConfiguration} for each step. 
% Specifically, for each start state, we sample 10 random interactions, one of which is the reversed actions, and record the corresponding ending states. 
The training of the \textbf{SpreadCloth} and \textbf{RopeConfiguration} needs more data, for the reason that, the states, kinematics and dynamics of these objects are much more complex.

From a starting state, we collect both successful interaction data using the proposed \textit{Fold to Unfold} data collection method, and failure interaction data using a random policy.
Therefore, the trained dense affordance could represent the distribution of diverse results of diverse actions.
Each interaction data contains the actions (picking point and placing point) and results after the action (\emph{e.g.} cloth coverage area for \textbf{SpreadCloth}).

\subsection{Hyper-parameters}
We set batch size to be 20, and use Adam Optimizer~\cite{kingma2014adam} with 0.0001 as the initial learning rate. 
During \textbf{Integrated Systematic Training (IST)} procedure,
we set learning rate to be 0.00005, as the  affordance modules have been trained before, and are only adapted and integrated into a system in this procedure.

    \subsection{Computing Resources}
We use TensorFlow as our Deep Learning framework. 
Each experiment is conducted on an RTX 3090 GPU, and consumes about 20 GB GPU Memory for training. It takes about 12 hours and 6 hours to respectively train the Placing Module and the Picking Module for one step. Besides, the Integrated Systematic Training procedure consumes 6 hours.

\section{Additional Details of Network Architectures}

The Picking Module and the Placing Module both employ Fully Convolutional Networks (FCNs) with the same structure to extract point-level features. Through the FCNs, the feature of the $W \times H \times C$ dimension input sequentially transforms to $W \times H \times 64$, $W \times H \times 64$, $W/2 \times H/2 \times 128$, $W/4 \times H/4 \times 256$, $W/8 \times H/8 \times 512$, $W/16 \times H/16 \times 512$ (bottleneck of the net work, where global feature is extracted), $W/8 \times H8 \times 512$, $W/8 \times H/8 \times 256$, $W/4 \times H/4 \times 256$, $W/4 \times H/4 \times 128$, $W/2 \times H/2 \times 128$, $W/2 \times H/2 \times 256$, $W \times H \times 256$, $W \times H \times 256$.

Afterwards, the Picking Module uses MLPs with hidden sizes to be (256$\to$256, 256$\to$1) to predict picking affordance, and the Picking Module uses MLPs with hidden sizes to be (1024$\to$256, 256$\to$1) to predict placing affordance. Here, 256 denotes the feature dimension of each (picking or placing) point, and 1024 denotes the dimension of the concatenation of the picking point feature (256), the placing point feature (256), and the global feature (512).

\section{Additional Details of Real-world Experiments}

\subsection{Real-robot Settings}

For real-robot experiments, 
we set up one Franka Panda robot on the workbench,
with a RealSense camera mounted on the robot gripper to take observations.
We use Robot Operating System (ROS)~\cite{quigley2009ros} to control the robot to execute actions.

\begin{figure}[htbp]
  \centering
  \includegraphics[width=\linewidth, trim={0cm, 0cm, 0cm, 0cm}, clip]{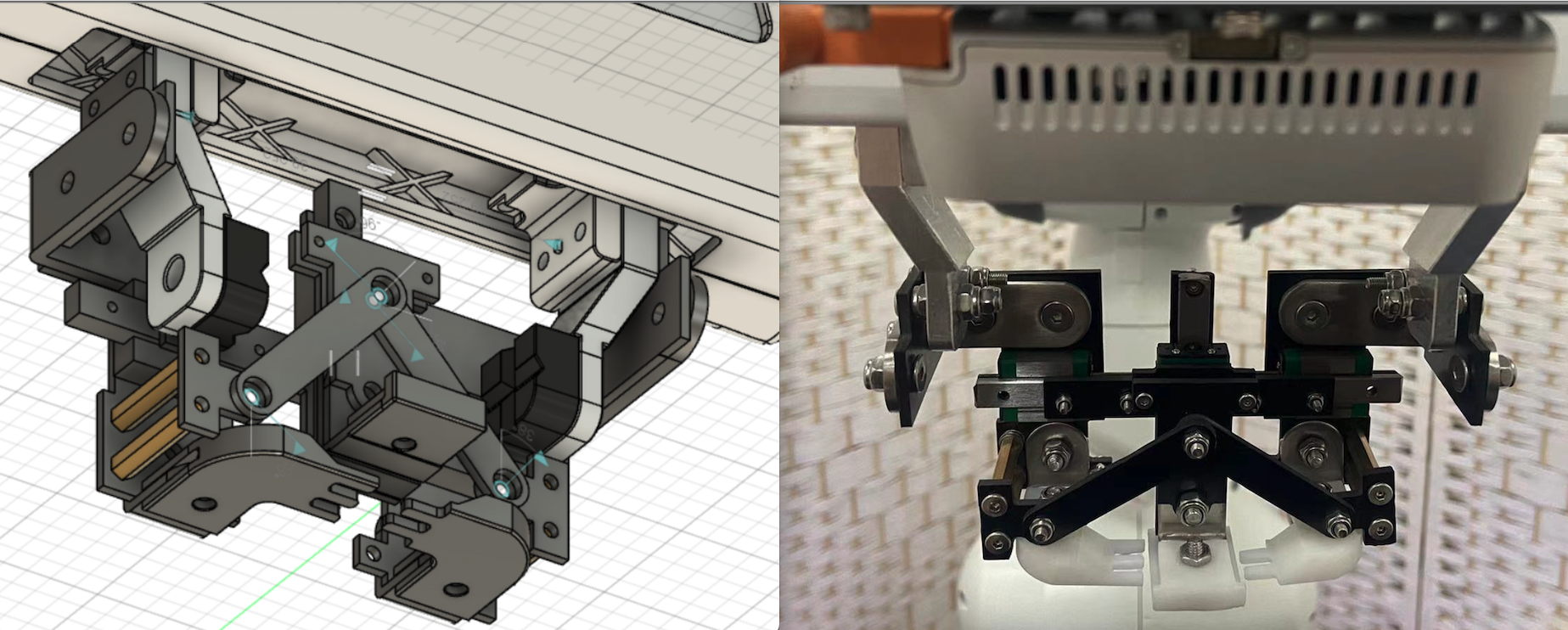}
  \caption{\textbf{Our Designed Fine-grained Fingers} to better pick deformable objects.}
  \label{fig:gripper}
\end{figure}

Additionally, as shown in Figure~\ref{fig:gripper}, as the original fingers of Franka Panda is wide and coarse,
to ensure that the gripper can pick only one layer of cloth instead of two layers at a time,
we design two fine-grained fingers mounted on the fingertips of the original fingers.

% \subsection{Domain Randomization}

\subsection{Real-world Data Collection and Fine-tuning}

As the configurations, kinematics and dynamics of deformable objects (like cloth and ropes) in the real world are different from those in simulation, we fine-tune the trained-in-simulation affordance using real-world collected interactions.

\begin{figure}[htbp]
  \centering
  \includegraphics[width=\linewidth, trim={0cm, 0cm, 0cm, 0cm}, clip]{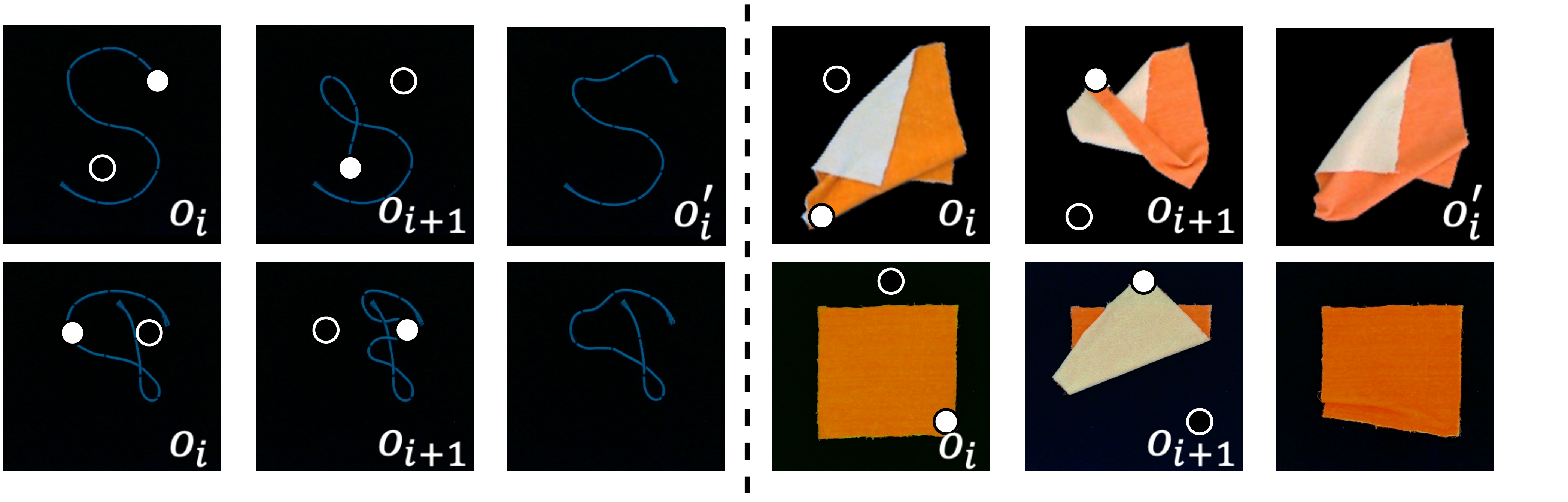}
  \caption{\textbf{Demonstrations of real-world collected data.} State $o_{i+1}$ shows the starting state and state $o_{i}^{\prime}$ show the ending state of interaction data for training.}
  \label{fig:real-world-data}
\end{figure}

Specifically, following Section 4.5 (\textbf{\textit{Fold to Unfold}: Efficient Multi-stage Data Collection for Learning Foresightful Affordance}) in the main paper,
we collect real-world interactions using the \textit{Fold-to-Unfold} method in different stages (demonstrations shown in Figure~\ref{fig:real-world-data}).
For fine-tuning,
we tune the learned picking and placing affordance stage-by-stage using the above real-world collected data.

\subsection{Metrics}

For \textbf{SpreadCloth}, we use the same metric as in~\ref{sec:simu_metric}, as it is easy to compute the coverage area of the cloth in the real world. 
For \textbf{RopeConfiguration}, we mark a black dot on the rope every 10cm, and compute the distances between these black dots and their ideal locations as the bipartite graph matching distance for further evaluation.

\subsection{Video Records of Real-world Manipulations}

Please see the {\color{red}\textbf{the videos in our project page}} for video records of real-world manipulations for both \textbf{SpreadCloth} and \textbf{RopeConfiguration} tasks.

\section{Assets}
We use the cable assets in \textit{DeformableRavens} benchmark as well as cloth and rope assets in \textit{SoftGym} benchmark, following their licenses. Our proposed assets with novel configurations can be generated using our code.

\end{document}